%% file: acl_latex.tex
\documentclass[11pt]{article}

\usepackage[preprint]{acl}

\usepackage{times}
\usepackage{latexsym}
\usepackage{enumitem}
\usepackage{amsmath}
\usepackage[utf8]{inputenc}
\usepackage[T1]{fontenc}
\usepackage{booktabs}
\usepackage{multirow}
\usepackage{tikz}
\usetikzlibrary{trees}
\usepackage{colortbl}
\usepackage{xcolor}
\usepackage{booktabs}
\usepackage{multirow}
\usepackage{subfigure}
\usepackage{fontawesome5}
\usepackage{mdframed}
\usepackage{xcolor}
\usepackage[T1]{fontenc}

\usepackage[utf8]{inputenc}

\usepackage{microtype}

\usepackage{inconsolata}

\usepackage{graphicx}
\usepackage[dvipsnames]{xcolor}
\usepackage{pifont}

\definecolor{impl}{RGB}{127,119,221}
\definecolor{deix}{RGB}{29,158,117}
\definecolor{spch}{RGB}{186,117,23}
\definecolor{socl}{RGB}{216,90,48}
\definecolor{cohr}{RGB}{55,138,221}

\definecolor{implbg}{RGB}{235,233,252}
\definecolor{deixbg}{RGB}{220,245,237}
\definecolor{spchbg}{RGB}{252,241,220}
\definecolor{soclbg}{RGB}{252,232,224}
\definecolor{cohrbg}{RGB}{220,235,252}

\title{VakyArth: Evaluating Pragmatic Competence in LLMs across \\ Indic Languages}

\author{
\textbf{Usneek Singh}$^{1*}$ \quad
\textbf{Poorvaja Veera Balaji Kumar}$^{1}$ \quad
\textbf{Parth Nanda}$^{1}$  \\
\textbf{Anand Madhusoodanan}$^{1}$ \quad
\textbf{Geyang Guo}$^{1}$ \quad
\textbf{Wei Xu}$^{1}$ \quad
\textbf{Junyi Jessy Li}$^{2}$ \\
\\
$^{1}$Georgia Institute of Technology, $^{2}$University of Texas at Austin \\
$^{*}$ \texttt{usingh68@gatech.edu} \\
}
\begin{document}
\maketitle

\begin{abstract}

    Real-world communication often requires pragmatic reasoning: interpreting meanings implied through context and cultural convention rather than stated literally. 
    Existing pragmatic evaluation remains largely limited to English and high-resource languages, leaving Indic languages unexplored despite their linguistic and cultural diversity. 
    We introduce \textbf{VakyArth}, the first pragmatic benchmark for Indic languages, designed as a diagnostic evaluation covering Hindi, Punjabi, Tamil, and Malayalam. 
    VakyArth evaluates models across five phenomena: deixis, speech acts, implicature, social pragmatics, and coherence; through multiple-choice questions, natural language inference, and translation, with all items authored by native speakers. 
    Across multilingual large language models (LLMs) of varying families and sizes, we find consistent failures on pragmatic meanings rooted in Indic linguistic and cultural conventions. 
    Our analysis shows systematic differences across languages and tasks: MCQ accuracy exceeds NLI accuracy in all model-language combinations, translation performance does not reliably track pragmatic understanding, and Indo-Aryan languages show a translation advantage over Dravidian languages. 
    We further show that automatic translation metrics can miss fluent but pragmatically unfaithful outputs, especially for implicature and deixis. 
\end{abstract}

\input{introduction}

\input{related_works}

\input{methodology}

\input{evaluation}
\input{results}
\input{conclusion}
\input{limitations}

\section*{Acknowledgments}
This work was partially funded by NSF Awards 2107524 and 2145479, as well as Good Systems,\footnote{\url{https://goodsystems.utexas.edu/}} a
UT Austin Grand Challenge to develop responsible AI technologies. We thank Kelly Marchisio for their valuable feedback. We also thank Cohere for providing API credits to support this work.

\bibliography{custom}
\newpage
\appendix

\input{appendix}
\end{document}

%% file: introduction.tex
\begin{figure}[t] 
    \centering
     \includegraphics[width=0.8\columnwidth]{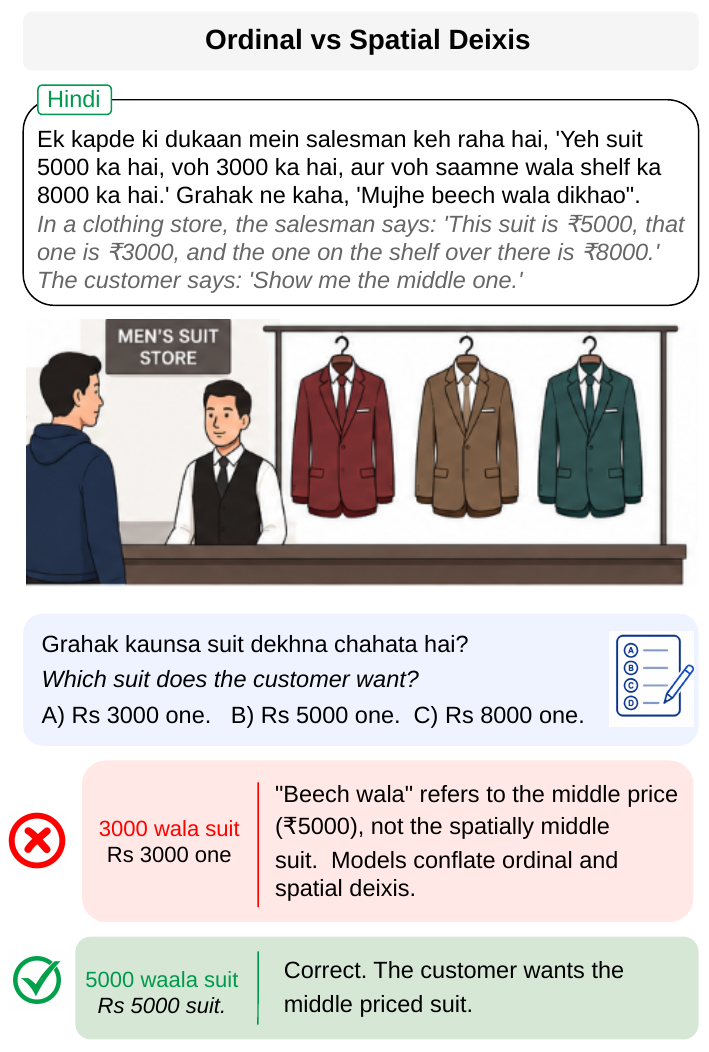}
    \caption{A Hindi deixis item from \textsc{VakyArth}. The customer asks for \textit{beech wala} (the middle one), referring to the middle-priced suit (Rs 5000). Any Hindi speaker resolves this by price ranking, not by the order of mention. Evaluated models pick Rs 3000 missing that \textit{beech wala} here anchors to price, not utterance order.}
    \label{fig:elements}
\end{figure}

\section{Introduction}

As large language models~(LLMs) become widely used, their impact depends not only on supporting professional tasks but also on helping users navigate everyday communication across diverse linguistic and cultural contexts.
India offers such a setting: it has the second-largest AI user base globally, yet usage remains concentrated in professional contexts such as software development, with much less adoption for everyday and personal tasks~\cite{appel2026indiacountrybrief}. 
Broadening AI's impact in India, therefore, requires moving beyond work-related applications toward systems that can support daily usage. 

Everyday communication in Indic contexts often depends on culturally grounded meanings that go beyond literal form. 
For example, as shown in Figure~\ref{fig:elements}, when a Hindi-speaking customer asks for ``\textit{beech wala}'' in a clothing-store dialogue, the phrase literally means ``\textit{the middle one}''. 
In context, however, the phrase refers to the middle-priced suit rather than the spatially middle item, a distinction missed by multiple existing LLMs (e.g., Gemma-4-31B). 
Pragmatic reasoning is needed here: recovering meaning from context, cultural convention, and speaker intent~\citep{ma2025pragmatics}.

Existing pragmatic evaluation mainly focuses on English or high-resource languages~\cite{ma2025pragmatics, park2024multiprageval}. 
To the best of our knowledge, no pragmatic evaluation exists for Indic languages. 
Furthermore, Indic pragmatics is highly diverse: languages across India span multiple families, scripts, and sociolinguistic communities, with pragmatic cues encoded differently.

We introduce \textbf{VakyArth}\footnote{\textit{VakyArth} combines \textit{Vakya} (utterance) and \textit{Arth} (meaning) from Sanskrit, reflecting the benchmark's focus on the meaning beneath the utterance.}, the first pragmatic benchmark for Indic languages. 
VakyArth is a diagnostic benchmark that covers Hindi, Punjabi, Tamil, and Malayalam, which represent diverse linguistic and regional contexts across India. 
It evaluates multilingual LLMs along five pragmatic phenomena: deixis, speech acts, implicature, social pragmatics, and coherence (see examples in Figure~\ref{fig:examples}). 
Each phenomenon is constructed in the corresponding language and evaluated under three task formats: multiple-choice questions, natural language inference, and translation, allowing us to assess pragmatic understanding at different levels of explicitness. 
All items are authored by native speakers and include naturally code-mixed instances to reflect real usage patterns.

We evaluate five multilingual LLMs across model families and sizes (ranging from 8B to 111B), and find consistent failure patterns rooted in Indic linguistic and cultural conventions. 
Models struggle with culturally grounded sarcasm and hyperbole, indirect refusals and requests, shifting deictic anchors such as ``kal'' and ``parso'', and social cues involving kinship address and gender norms.
Our quantitative analysis further reveals systematic differences across tasks, phenomena, and languages. 
Indo-Aryan languages show a translation advantage over Dravidian languages, likely reflecting differences in pretraining coverage~\cite{naous2025origin}. 
MCQ accuracy exceeds NLI accuracy in all model-language combinations, suggesting pragmatic interpretation benefits from answer-choice scaffolding.
Translation performance does not track MCQ or NLI scores, indicating that recognizing pragmatic meaning and expressing it faithfully across languages are distinct capabilities. Interestingly, Gemma-4-31B achieves the highest performance despite being smaller and lacking Indic specialization.
Automatic metrics also fail to penalize fluent but pragmatically unfaithful outputs for implicature and deixis.

We release our dataset and code at \url{https://github.com/Usneek1/VakyArth}.

%% file: related_works.tex
\section{Related Work}

\paragraph{Pragmatic evaluation of LLMs.}
The evaluation of pragmatic competence in LLMs has attracted growing attention. \citet{ma2025pragmatics} provide a comprehensive survey, noting that most resources cover a narrow set of phenomena and are heavily skewed toward high-resource languages. Early diagnostic work includes IMPPRES \citep{jeretic2020natural}, a targeted NLI dataset for implicature and presupposition, and \citet{ruis2023goldilocks}, who showed instruction-tuned models substantially outperform base models on implicature resolution. More comprehensive benchmarks have since emerged: PUB \citep{sravanthi2024pub} covers implicature, presupposition, reference, and deixis across fourteen multiple-choice tasks, and \citet{yu2026pragmatic} focus on alternative-based pragmatic reasoning. All of these benchmarks are in English.

\paragraph{Multilingual and cross-cultural pragmatic benchmarks.}
Extending pragmatic evaluation beyond English has been limited. MultiPragEval \citep{park2024multiprageval} evaluates models across English, German, Korean, and Chinese using Gricean maxims, but covers only European and East Asian languages. \citet{park2024pragmatic} and \citet{koo2025evaluating} introduce pragmatic benchmarks for Korean, with latter reporting substantial gaps between human and model performance on indirect speech acts. A related line of work evaluates cultural knowledge rather than pragmatic competence \citep{chiu2024culturalbench, shaikh2023modeling}. Unlike these studies, which test whether models know cultural facts, our work evaluates whether models can reason about implied meaning in context. No prior work evaluates pragmatic competence for any Indic language.

\paragraph{Evaluation of LLMs on Indic languages.}
Indic NLP evaluation has progressed from foundational benchmarks to increasingly culturally grounded ones, yet pragmatic competence remains unaddressed. Early work such as IndicNLPSuite \citep{kakwani2020indicnlpsuite} and IndicXTREME \citep{doddapaneni2023towards} established baselines for classification and named entity recognition. More recent work covers generation: IndicGenBench \citep{singh2024indicgenbench} evaluates translation and summarization across 29 languages, and MILU \citep{verma2025milu} introduces culturally specific knowledge tasks. The most culturally grounded effort is PARIKSHA \citep{watts2024pariksha}, which evaluates 30 models across 10 Indic languages and finds that LLM evaluators align poorly with humans on culturally nuanced responses — a finding that resonates with our own human-COMET divergence on pragmatic translation. Despite this progression, none of these benchmarks require pragmatic inference: understanding implied meaning, resolving deictic reference, or interpreting indirect speech acts. VakyArth is the first to address this gap.

%% file: methodology.tex
\newmdenv[
  topline=true, bottomline=true, rightline=true,
  leftline=true, linewidth=0.2pt, linecolor=black,
  backgroundcolor=yellow!5,
  innerleftmargin=5pt, innerrightmargin=5pt,
  innertopmargin=5pt, innerbottommargin=5pt,
  skipabove=5pt, skipbelow=5pt,
  roundcorner=5pt
]{phenbox}
\begin{figure*}[t] 
    \centering
     \includegraphics[width=\textwidth]{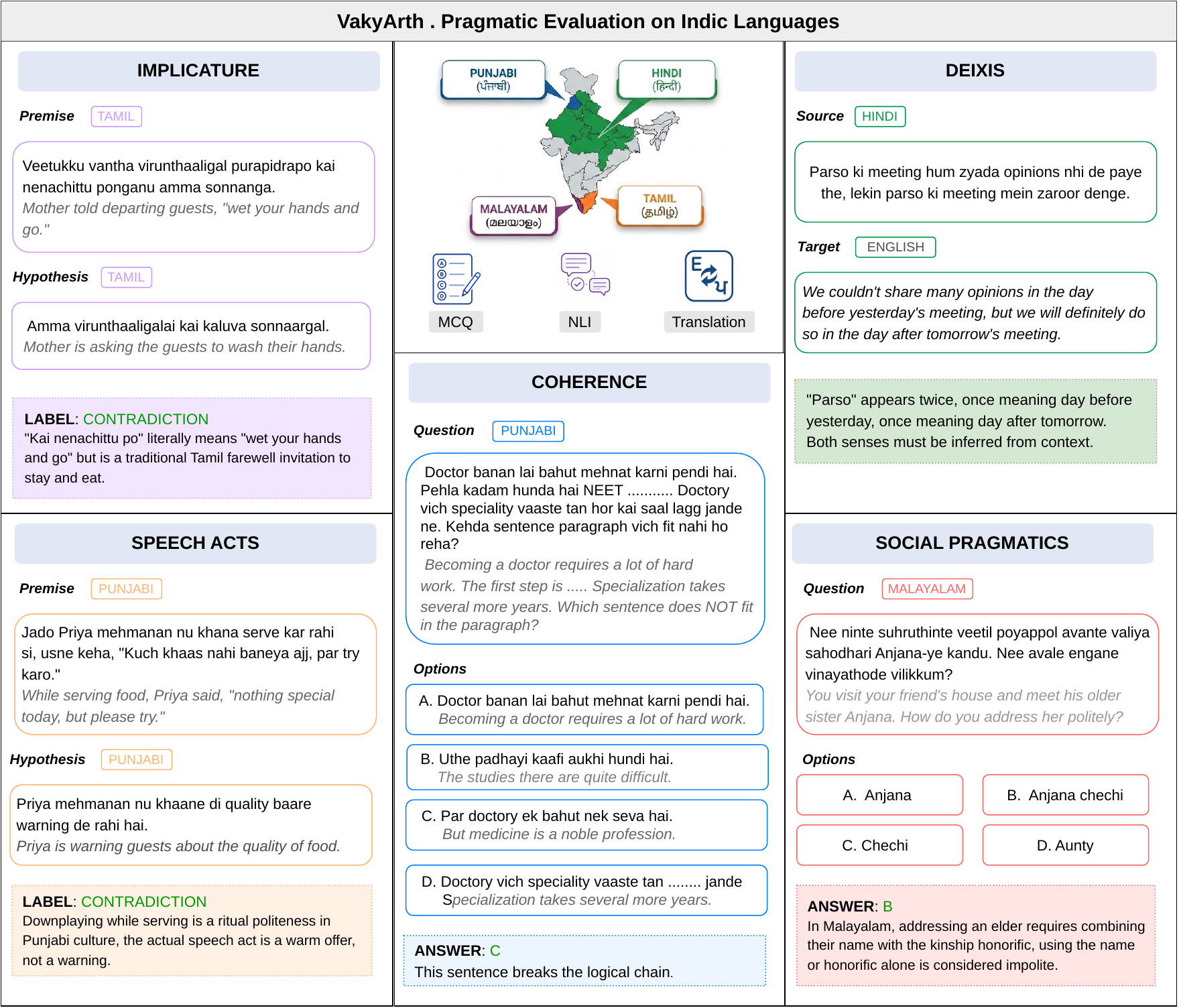}
    \caption{VakyArth benchmark overview. Items span five pragmatic phenomena : implicature, deixis, speech acts, social pragmatics, and coherence  across Hindi, Punjabi, Tamil, and Malayalam, evaluated under NLI, translation, and MCQ formats. In each case, the correct label or translation cannot be recovered from surface form alone.}
    \label{fig:examples}
\end{figure*}
\section{Evaluation Design}

 We design our evaluation around three guiding principles: \textbf{phenomena coverage} (the testing should span the core dimensions of pragmatic competence, not a single category), \textbf{task diversity} (evaluation should probe understanding at multiple levels of explicitness), and \textbf{cultural authenticity} (items should reflect how Indic languages are actually used, including code-mixing and culturally specific pragmatic conventions).
\subsection{Pragmatic Phenomena}
We evaluate five phenomena that collectively capture the range of non-literal, context-dependent meaning central to everyday communication (see Figure \ref{fig:examples}). These categories follow the taxonomy of \citet{ma2025pragmatics}, but are instantiated through distinctively Indic examples drawn from cultural knowledge of native speakers.

\vspace{0.7em}
\noindent\textbf{\colorbox{deixbg!100}{(1) Deixis}} refers to words whose reference depends entirely on the discourse
context, who is speaking, where, and when \citep{webber1988discourse}. Indic languages present particular deictic challenges: temporal words like \textit{kal} (yesterday/tomorrow) and \textit{parso} (day before yesterday/day after tomorrow) are bidirectional, resolving to past or future only through context; proximal-distal contrasts like \textit{intha-antha} carry meaning that translation flattens; and cultural calendar references. \textit{For example},

\begin{phenbox}
A Tamil item presents a speaker comparing two pumpkins: ``intha poosani paaka pinjaa irukku, antha poosani nallaa irukku'' (``this pumpkin looks less ripe, that one looks better'') illustrates how the proximal–distal contrast (\textit{intha} vs. \textit{antha}) carries the entire communicative point. A translation that flattens both to ``the pumpkin'' is grammatically fine but loses the comparison entirely.
\end{phenbox}

We also add multi-party dialogues for more complexity.
For example, in a Punjabi station exchange, speaker B says ``tussi edher aa jao'' (``you come over here'') while speaker A is already walking toward the wrong gate. Correctly resolving ``edher'' (``here'') requires tracking which gate each speaker is at - a reference chain that models often collapse.


\vspace{0.7em}
\noindent\textbf{\colorbox{spchbg!100}{(2) Speech Acts}} are utterances that perform actions such as requests, offers, refusals,
and warnings, whose illocutionary force often diverges substantially from
their surface form \citep{austin1975things}. South Asian communicative norms
make particularly heavy use of indirect speech acts as politeness strategies. \textit{For example,} 
\begin{phenbox}
A mother who enters her son's room at 11 AM and asks (in Hindi) \textit{"Beta, kya aaj chutti hai kya?"} ("Son, is it a holiday today?") is not seeking calendar information. She is delivering an indirect command to wake up and get to work. 
\end{phenbox}
Likewise, a Punjabi host who says ``\textit{kuch khaas nahi baneya}'' (``nothing special was made'') while placing food in front of guests is performing a ritual offer, not a sincere self-criticism.


\vspace{0.7em}
\noindent\textbf{\colorbox{implbg!100}{(3) Implicature}} refers to meaning conveyed but not explicitly stated, requiring the
listener to reason beyond the literal content \citep{bach2006speech}. In Indic languages, implicature is frequently
expressed through culturally embedded expressions that are opaque without
cultural knowledge. \textit{For example,}
 
\begin{phenbox}
A teacher who tells Rahul "Tumne toh iss pareeksha mein pass hokar kila jeet liya" ("You have conquered a fort by passing this exam") is using a common Hindi hyperbole to express praise. A model that processes the surface meaning predicts that Rahul won a real fort; the correct inference is that the teacher is simply congratulating him on passing.
\end{phenbox}
Similarly, in Tamil, a mother telling departing guests
\textit{``kai nenachu po''} (``wet your hands and go'') is not
issuing a hygiene instruction but extending a warm, face-saving
invitation to stay and eat. The ritual indirectness is the social
point of the expression.


\vspace{0.7em}
\noindent\textbf{\colorbox{soclbg!100}{(4) Social Pragmatics}} captures how social factors such as kinship, gender norms, power
dynamics, and cultural conventions around face shape language use
\citep{ma2025pragmatics}. These items require models to bring
cultural world knowledge to bear in ways that go beyond linguistic
competence. \textit{For example,}
\begin{phenbox}
In Punjabi, greeting a male friend with a hug while greeting his wife with folded hands (\textit{hath jodna}) and \textit{Sat Sri Akal} is the culturally appropriate norm for mixed-gender visits. A model without cultural grounding might incorrectly read this a sign of differential regard toward the wife.
\end{phenbox}
In Malayalam, addressing a friend's older sister requires combining
her name with the kinship honorific \textit{chechi}: ``Anjana chechi''
is the polite form, while ``Anjana'' alone signals disrespect and
``Chechi'' alone is too impersonal.


\noindent\textbf{\colorbox{cohrbg!100}{(5) Coherence}} requires reasoning about how utterances relate within a larger discourse
unit, beyond what any single sentence conveys \citep{van1997study}. 
Existing work draws extensively on sentence-ordering or narrative cloze tasks \cite{barzilay-lapata-2008-modeling,mostafazadeh-etal-2016-corpus,beyer2021incoherence}. In contrast,
our items test discourse-level reasoning within Indic language paragraphs, where topically plausible but discourse-incoherent distractors require models to track argumentative structure rather rely on keyword overlap. \textit{For example,} 
\begin{phenbox}
A Tamil paragraph describes a traditional banana leaf meal in sequence.. hot rice, lentils, ghee, kootu, poriyal ending with payasam. The most appropriate response the accompaniments like vadai and appalam that complete the feast. Options like "banana leaf meals are no longer common" or "it is difficult to serve on a leaf" are topically related to banana leaf meals but break the sequence the paragraph has built up.
\end{phenbox}

\subsection{Task Types}
Each phenomenon is evaluated across three task formats, chosen to probe pragmatic understanding at different levels of explicitness:

\vspace{0.2em}
\noindent\textbf{Multiple-Choice Question (MCQ).} The model picks the most appropriate interpretation from four options. We deliberately construct distractors that reflect the literal meaning of the utterance, so models that do not reason pragmatically will be misled.

\vspace{0.2em}
\noindent\textbf{Natural Language Inference (NLI).} The model labels the relationship between a premise and a hypothesis as entailment, contradiction, or neutral. 
This is intended to be harder than MCQ since no options to guide the model. We include more contradiction cases, especially for implicature, since models tend to treat figurative premises as entailing hypotheses that they actually contradict.

\vspace{0.2em}
\noindent\textbf{Translation.} The model translates an Indic-language utterance into English. Translating pragmatic content is challenging because implied meanings, cultural references, and speaker intent do not always have a direct equivalent in English. A model that does not understand the pragmatic meaning will produce a translation that is fluent but wrong. 

\subsection{Dataset Construction}
All items were written from scratch by a team of four native speakers, one per language, each with graduate-level NLP training. Annotators drew on personal linguistic knowledge, cultural familiarity, and grammar references to write items. Each item was reviewed by the full team to check for ambiguity, label correctness, and translation accuracy. Items where the correct answer was recoverable from surface cues alone were removed.
Most items are code-mixed with English. This reflects how people actually use these languages in daily life, especially in Hindi and Punjabi. Tamil and Malayalam items have less code-mixing, consistent with real usage patterns in those communities.

Table~\ref{tab:language_stats} summarizes the language statistics. VakyArth contains \textbf{578} items across four languages: Hindi (185), Punjabi (125), Tamil (134), Malayalam (134), spanning a combined speaker population exceeding 800 million. The corpus totals 14,921 words of source text and 424 human-authored English reference sentences in the translation subcorpus. The token/word figures reflect a fundamental typological difference between the two language families: Tamil and Malayalam are agglutinative, requiring significantly more tokens per word than the morphologically simpler Hindi \& Punjabi, consistently across all three tokenizers.

\begin{table}[t!]
\centering
\setlength{\tabcolsep}{1.5pt}
\renewcommand{\arraystretch}{1.15}
\footnotesize
\begin{tabular}{lcccc}
\toprule
& \textbf{Hindi} & \textbf{Punjabi} & \textbf{Tamil} & \textbf{Malay.} \\
\midrule
Family          & Indo-Aryan & Indo-Aryan & Dravidian & Dravidian \\
Script          & Devanagari & Gurmukhi   & Tamil     & Malayalam \\
\midrule
Items           & 185        & 125        & 134       & 134       \\
TR sen.    & 147        & 48         & 86        & 143       \\
\midrule
T/W (Llama)  & 1.95     & 1.90       & 3.15      & 4.44      \\
T/W (Gemma)  & 1.76     & 1.73       & 2.84      & 3.82      \\
T/W (Qwen)   & 1.96     & 1.87       & 3.13      & 4.32      \\
\bottomrule
\end{tabular}
\caption{Language statistics for VakyArth. T/W (Tokens/Words) is the average tokens
per word under each model's tokenizer for Latin-transliterated source text.
Dravidian languages consistently require 1.6--2.5$\times$ more tokens per
word than Indo-Aryan languages across all three tokenizers.
TR sentences = total sentences in the translation subcorpus.}
\label{tab:language_stats}
\end{table}




%% file: evaluation.tex
\section{Evaluation Setup} 

\paragraph{Models.} We evaluate five instruction-tuned models ranging from 8B to 111B parameters. Three are open-weight models run locally on an NVIDIA H100 GPU via the HuggingFace Transformers library \citep{wolf2020transformers}: \texttt{Llama-3-8B-Instruct}, \texttt{Gemma-4-31B}, and \texttt{Qwen3-32B}. The remaining two are open-weight multilingual models accessed via their respective APIs rather than run locally: \texttt{Sarvam-105B}, which has dedicated Indic language pretraining, and \texttt{Command-A-03-2025} (111B parameters), Cohere's flagship multilingual model. 
All models are run with a temperature of 0.2 and a maximum of 128 new tokens.

\paragraph{Prompting.} We use a three-shot prompting setup for all models and tasks. Each prompt contains three fixed examples written in the same language and script as the test item. Examples are shared across phenomena within a language but are task-specific. We use few-shot prompting rather than zero-shot because pilot runs showed that models frequently failed to follow the required output format under zero-shot conditions, making automatic parsing unreliable. More details on prompts can be found in Appendix \ref{sec:Appendix}.

\paragraph{Input format.} All items are presented in Latin transliteration. This choice reflects everyday usage: \citet{roark2020processing} documents informal romanization as the standard way South Asian languages are typed in digital communication, which is the setting our benchmark targets. Methodologically, \citet{jaavid2024romansetu} show that romanized Indic text reduces token fertility by 2--4$\times$ and matches or outperforms native-script representation across NLU, NLG, and MT tasks, supporting our choice to minimize tokenization artifacts. We additionally show the effect of using native script vs.\ Latin script on pragmatic understanding (see \S\ref{subsec: script}).

\paragraph{Evaluation metrics.} For MCQ and NLI, we report accuracy, computed by exact match between the model's predicted label and the gold answer. For translation, we report two scores: (1) automatic evaluation using COMET \citep{rei2020comet} with the \texttt{wmt22-comet-da} checkpoint, and (2) human evaluation scores on a 1--5 scale covering adequacy, fluency, and pragmatic adequacy, collected from four native speaker raters, one per language. We chose \texttt{wmt22-comet-da} over newer checkpoints after empirically validating that it achieves the highest correlation with human judgements in our setting compared to XCOMET \citep{guerreiro2024xcomet} and MetricX \citep{juraska2023metricx}; we present a detailed comparison of all three metrics in \S\ref{subsec:corelation}.

%% file: results.tex
\section{Results and Analysis} 
\label{sec:results}





\begin{table*}[t]
\centering
\setlength{\tabcolsep}{3.2pt}
\renewcommand{\arraystretch}{1.2}
\footnotesize

\textbf{Tasks:}\quad
\faClipboardList~MCQ (Accuracy)\quad
\faComments[regular]~NLI (Accuracy)\quad 
\faSync~Translation (COMET)

\vspace{3pt}

\begin{tabular}{ll
  ccc ccc ccc ccc ccc}
\toprule
& &
  \multicolumn{3}{c}{\cellcolor{deixbg}\textcolor{deix}{\textbf{Deixis}}} &
  \multicolumn{3}{c}{\cellcolor{spchbg}\textcolor{spch}{\textbf{Speech Acts}}} &
  \multicolumn{3}{c}{\cellcolor{implbg}\textcolor{impl}{\textbf{Implicature}}} &
  \multicolumn{3}{c}{\cellcolor{soclbg}\textcolor{socl}{\textbf{Social Prag.}}} &
  \multicolumn{3}{c}{\cellcolor{cohrbg}\textcolor{cohr}{\textbf{Coherence}}} \\[-2pt]
\cmidrule(lr){3-5}\cmidrule(lr){6-8}\cmidrule(lr){9-11}
\cmidrule(lr){12-14}\cmidrule(lr){15-17}
\textbf{Lang.} & \textbf{Model} &
  \faClipboardList & \faComments[regular] & \faSync &
  \faClipboardList & \faComments[regular] & \faSync &
  \faClipboardList & \faComments[regular] & \faSync &
  \faClipboardList & \faComments[regular] & \faSync &
  \faClipboardList & \faComments[regular] & \faSync \\
\midrule

\multirow{5}{*}{\textbf{Hindi}}
 & Llama-3-8B      & 0.64 & 0.38 & 0.80 & \textbf{1.00} & 0.45 & 0.78 & \underline{0.73} & 0.20 & 0.76 & \textbf{0.90} & 0.22 & 0.71 & 0.79 & 0.50 & 0.80 \\
 & Gemma-4-31B     & 0.79 & 0.46 & \textbf{0.85} & \textbf{1.00} & 0.45 & \textbf{0.84} & \textbf{0.91} & \underline{0.30} & \textbf{0.86} & \textbf{0.90} & 0.22 & \underline{0.82} & \underline{0.93} & 0.50 & \underline{0.83} \\
 & Qwen3-32B       & 0.71 & \underline{0.85} & 0.73 & \underline{0.90} & \textbf{0.73} & 0.54 & \underline{0.73} & \textbf{0.80} & 0.58 & \textbf{0.90} & \textbf{0.44} & 0.77 & \underline{0.93} & \textbf{0.92} & 0.76 \\
 & Sarvam-105B     & \textbf{0.93} & 0.69 & 0.81 & \textbf{1.00} & \underline{0.55} & 0.67 & \textbf{0.91} & \underline{0.30} & 0.71 & \textbf{0.90} & \underline{0.33} & 0.75 & \textbf{1.00} & \underline{0.75} & 0.77 \\
 & Command-A-111B  & \underline{0.86} & \textbf{0.92} & \underline{0.84} & \textbf{1.00} & \textbf{0.73} & \underline{0.84} & \textbf{0.91} & \textbf{0.80} & \underline{0.83} & \textbf{0.90} & \textbf{0.44} & \textbf{0.83} & \underline{0.93} & \textbf{0.92} & \textbf{0.83} \\
\midrule
\multirow{5}{*}{\textbf{Punjabi}}
 & Llama-3-8B      & 0.22 & 0.22 & 0.78 & 0.33 & \underline{0.44} & 0.75 & 0.10 & \underline{0.33} & 0.67 & 0.67 & \underline{0.11} & 0.61 & 0.40 & \textbf{0.50} & 0.86 \\
 & Gemma-4-31B     & \textbf{0.78} & \underline{0.33} & \textbf{0.85} & \textbf{1.00} & \textbf{0.56} & \textbf{0.82} & \textbf{0.60} & \underline{0.33} & \textbf{0.78} & \textbf{1.00} & \textbf{0.22} & \textbf{0.79} & \underline{0.60} & \underline{0.33} & 0.89 \\
 & Qwen3-32B       & 0.56 & 0.11 & 0.58 & \underline{0.78} & 0.11 & 0.49 & \underline{0.50} & \underline{0.33} & 0.35 & \underline{0.83} & \textbf{0.22} & 0.43 & 0.40 & \underline{0.33} & 0.52 \\
 & Sarvam-105B     & \underline{0.67} & \textbf{0.44} & 0.84 & 0.67 & 0.33 & \underline{0.80} & \underline{0.50} & \underline{0.33} & 0.55 & \underline{0.83} & \textbf{0.22} & \underline{0.71} & \textbf{0.80} & \textbf{0.50} & \underline{0.89} \\
 & Command-A-111B  & \underline{0.67} & \textbf{0.44} & \underline{0.85} & \underline{0.78} & 0.22 & 0.80 & 0.40 & \textbf{0.44} & \underline{0.73} & \underline{0.83} & \textbf{0.22} & 0.68 & \textbf{0.80} & \textbf{0.50} & \textbf{0.89} \\
\midrule
\multirow{5}{*}{\textbf{Tamil}}
 & Llama-3-8B      & 0.33 & \underline{0.38} & 0.58 & \underline{0.44} & 0.33 & 0.54 & \underline{0.67} & 0.33 & 0.53 & \underline{0.67} & \underline{0.33} & 0.50 & 0.11 & \underline{0.56} & 0.53 \\
 & Gemma-4-31B     & \textbf{0.89} & \textbf{0.50} & \textbf{0.79} & \textbf{0.89} & \textbf{0.67} & \underline{0.82} & \textbf{0.78} & \underline{0.44} & \underline{0.71} & \textbf{0.78} & \textbf{0.56} & \textbf{0.77} & \underline{0.56} & \textbf{0.78} & \underline{0.76} \\
 & Qwen3-32B       & \underline{0.67} & 0.25 & 0.59 & 0.33 & \underline{0.44} & 0.57 & \underline{0.67} & \textbf{0.56} & 0.47 & 0.56 & 0.22 & 0.52 & \textbf{0.67} & \textbf{0.78} & 0.51 \\
 & Sarvam-105B     & \underline{0.67} & 0.25 & \underline{0.79} & \underline{0.44} & \underline{0.44} & 0.75 & \textbf{0.78} & 0.00 & \textbf{0.82} & \textbf{0.78} & \underline{0.33} & \underline{0.75} & \underline{0.56} & \underline{0.56} & 0.66 \\
 & Command-A-111B  & 0.33 & 0.12 & 0.78 & \underline{0.44} & \underline{0.44} & \textbf{0.82} & \underline{0.67} & 0.22 & 0.69 & 0.44 & 0.22 & 0.67 & \underline{0.56} & \textbf{0.78} & \textbf{0.76} \\
\midrule
\multirow{5}{*}{\textbf{Malayalam}}
 & Llama-3-8B      & 0.22 & 0.22 & 0.58 & 0.56 & \textbf{0.56} & 0.61 & 0.50 & \textbf{0.22} & 0.55 & 0.44 & 0.22 & 0.58 & 0.56 & 0.44 & 0.59 \\
 & Gemma-4-31B     & \textbf{0.89} & \underline{0.44} & \textbf{0.83} & \textbf{1.00} & \textbf{0.56} & \textbf{0.80} & \textbf{1.00} & \textbf{0.22} & \textbf{0.78} & \textbf{1.00} & 0.22 & \textbf{0.78} & \textbf{1.00} & 0.56 & \textbf{0.75} \\
 & Qwen3-32B       & 0.33 & \underline{0.44} & 0.55 & 0.78 & \textbf{0.56} & 0.61 & 0.62 & \textbf{0.22} & 0.56 & \underline{0.78} & \underline{0.33} & 0.61 & 0.78 & \underline{0.67} & 0.65 \\
 & Sarvam-105B     & \underline{0.67} & \textbf{0.67} & 0.37 & 0.78 & \textbf{0.56} & 0.38 & \underline{0.88} & \textbf{0.22} & 0.35 & \underline{0.78} & \textbf{0.44} & 0.36 & \underline{0.89} & \textbf{0.78} & 0.36 \\
 & Command-A-111B  & 0.56 & \underline{0.44} & \underline{0.77} & \underline{0.89} & \textbf{0.56} & \underline{0.77} & \underline{0.88} & \underline{0.11} & \underline{0.75} & \textbf{1.00} & 0.22 & \underline{0.74} & \underline{0.89} & 0.44 & \underline{0.74} \\

\bottomrule
\end{tabular}

\vspace{3pt}
\footnotesize Best score per column in \textbf{bold}; second best is \underline{underlined}.

\caption{MCQ accuracy, NLI accuracy, and translation COMET scores for all five models across four languages and five pragmatic phenomena. MCQ accuracy exceeds NLI accuracy in every model-language combination. Best score per column in bold; second-best underlined.}
\label{tab:main_results}
\end{table*}

Table~\ref{tab:main_results} presents results across all four models, four languages, and five phenomena for MCQ accuracy, NLI accuracy, and COMET score. We organize our analysis around the pragmatic failure patterns that VakyArth uncovers. We discuss what is specifically challenging about Indic pragmatics for current models. We also present some quantitative patterns and insights in \S\ref{sec:quant}.

\subsection{Pragmatic Failure Patterns}
\label{sec:failures}
\paragraph{Deictic Reference Failure.}
Models frequently struggle to resolve deictic expressions that shift fluidly across discourse contexts. In particular, models make mistakes on bidirectional temporal expressions in Hindi and Punjabi. For instance, if a speaker says on Monday, \textit{``kal mera interview hai''} (``My interview is tomorrow''), and is asked on Tuesday how it went, Llama, Command-A and Gemma fail to correctly anchor the event timeline. Beyond absolute shifts, models also fail on relative calendar references conventionalized among native speakers: when presented with a Friday utterance like \textit{``agle hafte Somvar ko exam hai''} (``The exam is next week on Monday''), models compute the target date as seven days from the utterance rather than three.

\paragraph{Culturally Embedded Expressions.}
Models make mistakes in understanding idiomatic expressions rooted in regional socio-religious traditions. For instance, when a mother in Tamil tells departing guests \textit{``kai nenachittu ponganu''} (``wet your hands and go''), Llama, Gemma, and Qwen predict an instruction to return home and eat, while Sarvam and Command-A classify it as a hygiene command. No model recovers the correct pragmatic meaning which is a conventionalized invitation to stay for a meal. Similarly, when someone in Punjabi states \textit{``bas meri kudi da viyah karke mai ganga nha lavan''} (``once my daughter is married, I will bathe in the Ganga''), all models predict a literal travel intention, missing the culturally embedded vow signifying relief at fulfilling a major parental duty.

\paragraph{Indirect Speech Acts and Soft Refusals.}
South Asian communicative norms strongly disprefer direct refusals and favor indirect speech acts that models misinterpret at face value. For instance, when a speaker declines an invitation with the standard Hindi hedge \textit{``dekhte hain, try karungi''} (``Let's see, I'll try''), models predict she intends to attend. Native speakers recognize this as a polite soft declination. When a Tamil host asks a departing guest \textit{``enna avasaram''} (``What is the rush?''), models interpret it as an inquiry into an actual emergency rather than a standard hospitality request, urging the guest to stay longer.

\paragraph{Social and Kinship Norms.}
Models fail to interpret utterances whose meaning depends on regional social conventions, or ritual politeness scripts. For instance, in Punjabi culture, an elder telling a younger relative \textit{"tu taan bada kamzor lag reha hai"} (You look quite thin) is a conventionalized expression of care; all five models misread it as mockery. Models also fail to parse inverted politeness scripts: when a young relative accepts cash at a Malayalam wedding saying \textit{"ithonnum vendaayirunnu, chetta/chechi"} (This wasn't necessary, brother/sister), models predict a genuine rejection, failing to recognize the socially obligatory script of verbal refusal during physical acceptance. The ritual farewell \textit{"poyittu varam"} (I will go and come back) is read as a literal promise of immediate return by the evaluated models.

\paragraph{Sarcasm, Hyperbole, and Implied Criticism.}
Models misinterpret non-literal assertions, such as detecting hyperbole or decoding sarcasm used for implied criticism. For example, when a character arrives late and their aunt remarks in Punjabi, \textit{``VIP aa gaye, hun program shuru kar dinde aa''} (``The VIP has arrived, now we can start the program''), all five models interpret the utterance as a genuine warm welcome. A native speaker instantly recognizes the pointed sarcasm about the lateness. Similarly, when a father in Malayalam exclaims \textit{``ee veettil enikku shwaasam polum kittunnilla''} (``I cannot even breathe in this house''), all models predict literal respiratory distress, missing the culturally transparent hyperbolic expression of frustration.

\subsection{Model and Task comparisons}
\label{sec:quant}
\paragraph{Which model performs best?}
Gemma-4-31B is the strongest model overall in the benchmark (MCQ 0.86 NLI 0.43 COMET 0.81). Command-A-111B is the next strongest model, close behind Gemma on MCQ (0.74) and COMET (0.78), and edging ahead of it on NLI (0.46 vs.\ 0.43). Sarvam-105B is comparable to Command-A on MCQ (0.77) but noticeably weaker on translation (COMET 0.66). Despite being smaller than Sarvam-105B and Command-A-111B and lacking dedicated Indic pretraining, Gemma-4-31B is the most reliable model in our evaluation.
\paragraph{Do all tasks have equal performance?}
MCQ scores are higher than NLI scores in all 20 of 20 model-language aggregates. This mirrors the distinction between receptive and productive linguistic competence in psycholinguistics: recognizing a correct interpretation among options is cognitively easier than generating or inferring it without scaffolding \citep{ruis2023goldilocks}.
\paragraph{Do all languages have the same performance?}
The Dravidian languages, Tamil and Malayalam, lag behind on translation (though this gap is not consistent across MCQ and NLI) than the Indo-Aryan languages, Hindi and Punjabi. This is likely because of lower Dravidian language representation in model pretraining data rather than any structural difference in pragmatic difficulty. The translation gap is most pronounced for Coherence and Deixis, and smallest for Implicature.

\subsection{Error patterns}
\begin{figure}
    \centering
    \includegraphics[width=\columnwidth]{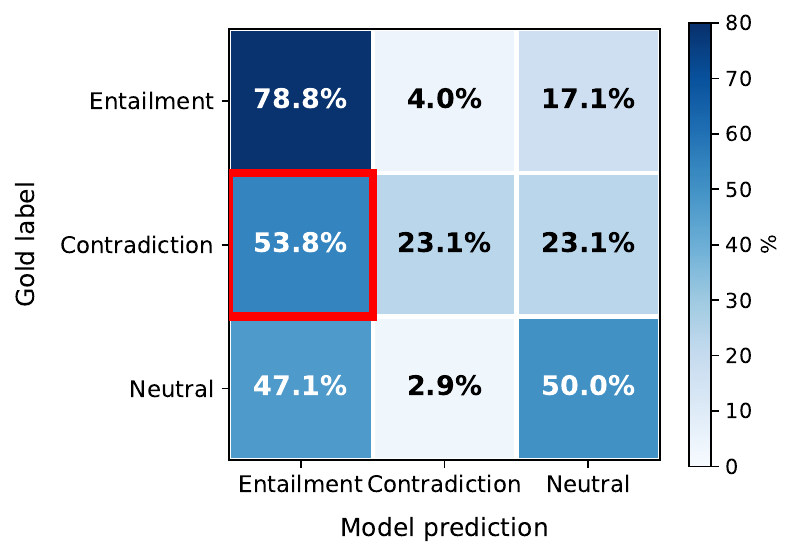}
    \caption{NLI confusion matrix pooled across all four models. Rows are gold labels, columns are model predictions. The highlighted cell marks the systematic bias toward Entailment (literal meaning in our case) when correct label is Contradiction.}
    \label{fig:conf_matrix}
\end{figure}

In this section we look inside the errors, examining whether models fail in predictable directions, whether failures are shared across model families, and whether NLI errors reveal systematic biases.
\paragraph{Do models default to literal meaning?}
A design choice in VakyArth was to construct MCQ distractors that reflect the literal, surface reading of each utterance. If models are failing because they cannot reason pragmatically, they should fall into these traps. We test this directly on the MCQ subset where a literal-meaning distractor is unambiguously identifiable, and look at what models pick when they get the item wrong. Across all wrong-answer events, models chose the literal distractor \textbf{59.1\%} of the time, nearly twice the 33.3\% rate expected by chance \((p < 0.001)\). This shows models are failing in a specific and predictable direction. For downstream applications, this means pragmatic failures are systematic and predictable rather than noisy, and can potentially be mitigated through targeted prompting or training.
\paragraph{Are failures model-specific or shared across models?}
For each item, we counted how many of the four models failed. If failures were driven by model-specific weaknesses, we would expect them scattered across different items. Instead, models fail on the same items. On NLI, \textbf{24.7\%} of items are failed by all four models, rising to 41.7\% for Social Pragmatics. This suggests that the items where all models fail together isolate cases where pragmatic reasoning breaks down regardless of the model's overall capability.
\paragraph{Do models show systematic label bias on NLI?}
We construct our NLI dataset more as contradiction labels (57.5\%) because pragmatic reasoning often involves recognizing that a figurative or indirect utterance contradicts its literal reading. If models default to the literal reading, they should systematically mislabel these items as Entailment. Figure \ref{fig:conf_matrix} shows this is exactly what happens. When the gold label is Entailment, models get it right 78.8\% of the time. But when the gold label is Contradiction, models predict Entailment 53.8\% of the time — more often than they predict Contradiction (23.1\%). 
The bias also compounds our earlier finding on literal distractors: when scaffolding is removed, models do not merely guess randomly among labels — they systematically default to the reading where the utterance means what it literally says.

\subsection{Human Evaluation and COMET Correlation} \label{subsec:corelation}
We collect human translation scores on a 1--5 scale covering adequacy, fluency, and pragmatic adequacy from four native raters, one per language. Given the cost of native-speaker annotation, we sample a subset of 104 items (Hindi 31, Punjabi 26, Tamil 23, Malayalam 24) out of 204 total items, translated by four of the five evaluated models (Llama-3-8B, Gemma-4-31B, Qwen3-32B, Sarvam-105B), producing $4 \times 104 = 416$ scored outputs across all four languages and five phenomena. We compare them against three automatic metrics: COMET \citep{rei2020comet} with the \texttt{wmt22-comet-da}
checkpoint, XCOMET \citep{guerreiro2024xcomet}, and MetricX
\citep{juraska2023metricx}.
 
As shown in Table~\ref{tab:correlation}, COMET achieves the highest overall
correlation with human judgements (Pearson $r=0.813$, Spearman $\rho=0.860$),
outperforming both XCOMET ($\rho=0.792$) and MetricX ($\rho=0.788$) despite
being an older checkpoint. We hypothesize that XCOMET and MetricX, optimized
on formal translation benchmarks, generalize less well to the colloquial,
code-mixed, and culturally embedded text in our benchmark.
Correlation varies across languages and phenomena. All three metrics agree most with human scores on Coherence and Social Pragmatics, where translation errors tend to be large and obvious. Agreement is weakest on Implicature and Deixis, where models produce fluent translations that sound correct but miss the pragmatic meaning, the kind of subtle error that automatic metrics struggle to detect. This finding has a practical implication: automatic evaluation alone is insufficient for benchmarks where pragmatic faithfulness and surface fluency can diverge, and human judgment remains necessary for phenomena like Implicature and Deixis. We show a few such examples in \S\ref{subsec:divergence}.



\begin{table}[h]
\centering
\setlength{\tabcolsep}{4pt}
\renewcommand{\arraystretch}{1.2}
\small
\begin{tabular}{lcccccc}
\toprule
& \multicolumn{2}{c}{\textbf{COMET}}
& \multicolumn{2}{c}{\textbf{XCOMET}}
& \multicolumn{2}{c}{\textbf{MetricX}} \\
\cmidrule(lr){2-3}\cmidrule(lr){4-5}\cmidrule(lr){6-7}
& $r$ & $\rho$ & $r$ & $\rho$ & $r$ & $\rho$ \\
\midrule
All          & \textbf{0.81} & \textbf{0.86} & 0.80 & 0.79 & 0.73 & 0.79 \\
\midrule
\multicolumn{7}{l}{\textit{By language}} \\
Hindi        & \textbf{0.85} & \textbf{0.77} & 0.65 & 0.54 & 0.60 & 0.60 \\
Punjabi      & 0.77 & \textbf{0.80} & \textbf{0.78} & 0.77 & 0.74 & 0.76 \\
Tamil        & \textbf{0.82} & \textbf{0.80} & 0.72 & 0.61 & 0.62 & 0.62 \\
Malayalam    & 0.68 & \textbf{0.81} & \textbf{0.85} & 0.76 & 0.71 & 0.73 \\
\midrule
\multicolumn{7}{l}{\textit{By phenomenon}} \\
Implicature  & 0.81 & 0.83 & \textbf{0.80} & \textbf{0.82} & 0.69 & 0.75 \\
Deixis       & 0.80 & 0.81 & \textbf{0.85} & \textbf{0.81} & 0.72 & \textbf{0.82} \\
Speech Acts  & 0.76 & \textbf{0.87} & 0.77 & 0.79 & \textbf{0.74} & 0.80 \\
Social Prag. & \textbf{0.86} & \textbf{0.87} & 0.85 & \textbf{0.86} & 0.77 & 0.83 \\
Coherence    & \textbf{0.87} & \textbf{0.88} & 0.85 & 0.82 & 0.78 & 0.82 \\
\bottomrule
\end{tabular}
\caption{Pearson ($r$) and Spearman ($\rho$) correlation of three automatic
metrics with human judgements across all four languages and five phenomena
($n=416$). COMET achieves the highest overall correlation ($\rho=0.86$)
despite being an older checkpoint compared to MetricX.
XCOMET is weaker on Hindi
and Tamil.}
\label{tab:correlation}
\end{table}
  

\subsection{Script Experiment}
We evaluate all MCQ items across all five phenomena, in both native script (Devanagari, Gurmukhi, Tamil, and Malayalam) and Latin transliteration, to study whether script choice affects pragmatic understanding. Pooled across all languages, phenomena, and models, native script yields a modest overall advantage (76.5\% vs.\ 70.6\% accuracy). However, the effect is neither universal nor uniform in size: Punjabi (Qwen3-32B) and Malayalam (Command-A) both perform slightly worse in native script, and the size of the advantage where it does appear varies widely, from negligible (Hindi, Command-A: no difference) to substantial (Malayalam, Qwen3-32B: +23.0 points; Tamil, Command-A: +17.8 points). No single model is uniformly script-invariant or uniformly script-sensitive across languages, and no consistent language-level pattern emerges either. We leave a deeper investigation of this effect to future work. Since VakyArth targets everyday communication, we adopt Latin transliteration as the standard script in our evaluation.

\label{subsec: script}
\begin{figure}[h]
  \centering
  \includegraphics[width=0.48\columnwidth]{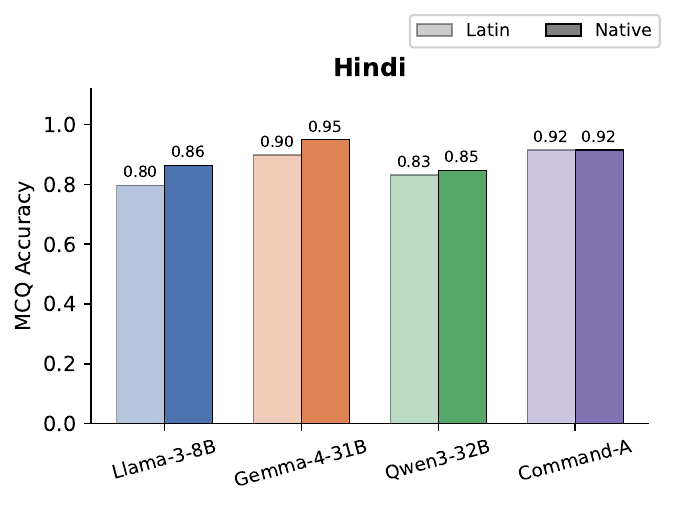}
  \hfill
  \includegraphics[width=0.48\columnwidth]{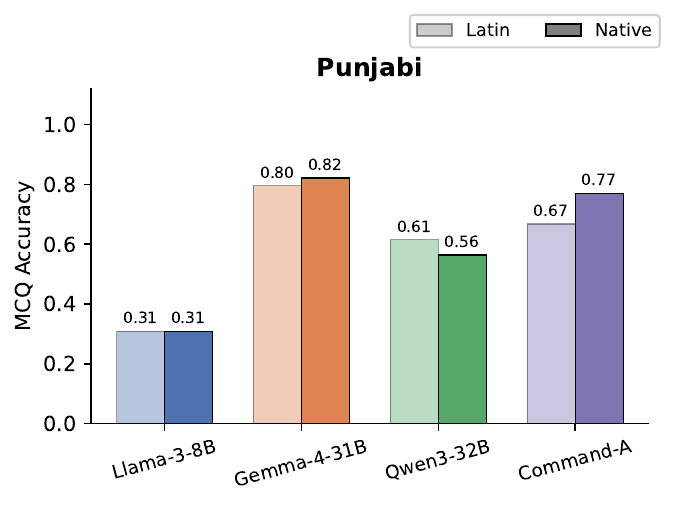}

  \vspace{4pt}

  \includegraphics[width=0.48\columnwidth]{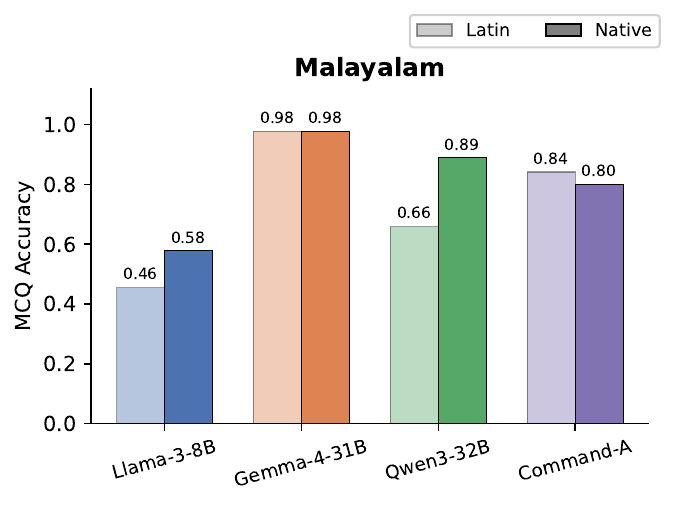}
  \hfill
  \includegraphics[width=0.48\columnwidth]{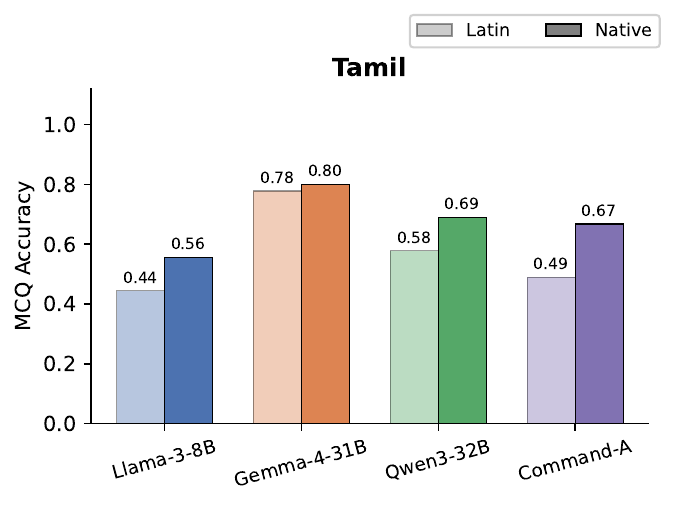}

  \caption{Script experiment: native script vs.\ Latin transliteration MCQ accuracy for all four languages, pooled across all five phenomena. No single model is uniformly
script-invariant or uniformly script-sensitive across languages}
  \label{fig:script}
\end{figure}

%% file: conclusion.tex
\section{Conclusion}

We present \textbf{VakyArth}, the first diagnostic benchmark for pragmatic competence in Indic languages, covering Hindi, Punjabi, Tamil, and Malayalam across five phenomena and three task types. Evaluating five models ranging from 8B to 111B parameters, we find consistent, phenomenon-specific failure patterns that existing benchmarks cannot detect. Gemma-4-31B is the strongest model overall despite lacking Indic specialization, and a persistent MCQ-NLI gap suggests that recognizing and reasoning about pragmatic meaning are distinct competencies. When models fail, they do so in a systematic and predictable direction defaulting to literal readings rather than guessing randomly among labels. Standard automatic metrics, while broadly reliable, fail to penalize fluent but pragmatically unfaithful outputs for implicature and deixis. We hope VakyArth serves as a foundation for building more pragmatically capable and culturally aware language technologies for Indic language users.

%% file: limitations.tex
\section{Limitations}
VakyArth has three main limitations. First, the script experiment covers only MCQ items; a fuller ablation and detailed investigation across all models, tasks, and languages remains future work. Second, all items are single-turn utterances or short passages and do not cover multi-turn pragmatics, which is an important dimension of real-world language use. Third, while our four languages represent a diverse sample of Indic language families, hundreds of other Indic languages remain uncovered.

%% file: appendix.tex
\section{Appendix}
\label{sec:Appendix}

\subsection{Prompt Templates}
\label{subsec:prompts}

We use 3-shot prompts for all models and tasks. Instructions and
few-shot examples are written in the source language to avoid
introducing English-language bias into the evaluation. The prompt
structure is identical across all four languages, only the
instruction text and examples are adapted to each language by a
native speaker. Below we show the templates for each task type
using Hindi as a representative example.
\subsection*{Translation Prompt}
 
\begin{mdframed}[
  topline=true, bottomline=true, rightline=true, leftline=true,
  linewidth=0.4pt, linecolor=black,
  backgroundcolor=gray!4,
  innerleftmargin=8pt, innerrightmargin=8pt,
  innertopmargin=6pt, innerbottommargin=6pt,
  roundcorner=2pt
]
\ttfamily\small
Neeche diye gaye Hindi text ko English mein anuvad karo.\\
Sirf translated text hi output mein dikhayein.\\[6pt]
 
Source:\\
Unhone milkar kaam kiya.\\[4pt]
Translation:\\
They worked together.\\[6pt]
 
Source:\\
Bazaar mein bahut saamaan milta hai.\\[4pt]
Translation:\\
A wide variety of goods is available in the market.\\[6pt]
 
Source:\\
Usne apna vaada nibhaya aur samay par pahunch gaya.\\[4pt]
Translation:\\
He kept his promise and arrived on time.\\[6pt]
 
Source:\\
\textit{[source utterance]}\\[4pt]
Translation:
\end{mdframed}
\newpage
\subsection*{MCQ Prompt}
 
\begin{mdframed}[
  topline=true, bottomline=true, rightline=true, leftline=true,
  linewidth=0.4pt, linecolor=black,
  backgroundcolor=gray!4,
  innerleftmargin=8pt, innerrightmargin=8pt,
  innertopmargin=6pt, innerbottommargin=6pt,
  roundcorner=2pt
]
\ttfamily\small
Is prashan ka sahi uttar kya hai?\\
Jabab mein sirf option ka akshar (A/B/C...) aur us option ka text
likho.\\[6pt]
 
Context:\\
Priya ne kaha, ``Aaj khana bahut achha bana hai.''\\[4pt]
Options:\\
A. Priya ko khana pasand aaya\\
B. Priya ne khana nahi khaya\\
C. Priya ne khana banaya\\
D. Priya ko bhook nahi thi\\[4pt]
Answer:\\
A. Priya ko khana pasand aaya\\[6pt]
 
Context:\\
Ramesh ne kaha, ``Bahar bahut thand hai.'' Lekin use bahar jaana tha.\\[4pt]
Options:\\
A. Ramesh bahar nahi jayega\\
B. Ramesh bahar jayega\\
C. Ramesh ko thand nahi lagti\\
D. Ramesh ghar mein rehna chahta hai\\[4pt]
Answer:\\
B. Ramesh bahar jayega\\[6pt]
 
Context:\\
Maa ne kaha, ``Koi baat nahi, agli baar dhyan rakhna.''\\[4pt]
Options:\\
A. Maa gusse mein hai\\
B. Maa ne maafi maangi\\
C. Maa ne tasalli di\\
D. Maa ne saza di\\[4pt]
Answer:\\
C. Maa ne tasalli di\\[6pt]
 
Context:\\
\textit{[source utterance]}\\[4pt]
Question:\\
\textit{[question]}\\[4pt]
Options:\\
\textit{[A. option 1]}\\
\textit{[B. option 2]}\\
\textit{[C. option 3]}\\
\textit{[D. option 4]}\\[4pt]
Answer:
\end{mdframed}
 
\newpage
\subsection*{NLI Prompt}
 
\begin{mdframed}[
  topline=true, bottomline=true, rightline=true, leftline=true,
  linewidth=0.4pt, linecolor=black,
  backgroundcolor=gray!4,
  innerleftmargin=8pt, innerrightmargin=8pt,
  innertopmargin=6pt, innerbottommargin=6pt,
  roundcorner=2pt
]
\ttfamily\small
Neeche diye gaye do vaakyon ke beech sambandh batao.\\
Jabab mein sirf `Entailment', `Contradiction', ya `Neutral' likho.\\[6pt]
 
Premise:\\
A: Khana taiyaar hai?\\
B: Main abhi bazaar gaya tha.\\[4pt]
Hypothesis:\\
Khana taiyaar hai.\\[4pt]
Label:\\
Contradiction\\[6pt]
 
Premise:\\
A: Kya tujhe pata hai kal chutti hai?\\
B: Haan, main subah se plan kar raha hoon.\\[4pt]
Hypothesis:\\
B kal kuch karna chahta hai.\\[4pt]
Label:\\
Entailment\\[6pt]
 
Premise:\\
A: Bahar baarish ho rahi hai.\\
B: Theek hai.\\[4pt]
Hypothesis:\\
B bahar jayega.\\[4pt]
Label:\\
Neutral\\[6pt]
 
Premise:\\
\textit{[source premise]}\\[4pt]
Hypothesis:\\
\textit{[hypothesis]}\\[4pt]
Label:
\end{mdframed}
 

\subsection{Annotation Guidelines for Translation Rating}
\label{subsec:annotation}

Human evaluation scores for translation outputs were collected from
four native speaker annotators, one per language, each with graduate-level
NLP training. Annotators rated each translation on a 1--5 scale assessing
how well the model preserved the \textit{pragmatic intent} of the source
utterance including implied meaning, tone, cultural nuance, and
illocutionary force rather than surface-level adequacy or fluency alone.
Annotators were instructed to read the source utterance, its English gloss,
and the model output, and assign a score based on the following rubric.

\begin{table}[h]
\centering
\setlength{\tabcolsep}{6pt}
\renewcommand{\arraystretch}{1.3}
\small
\begin{tabular}{cp{0.78\columnwidth}}
\toprule
\textbf{Score} & \textbf{Criterion} \\
\midrule
5 & Pragmatic intent fully preserved, including tone, implicature,
    and cultural nuance. \\
4 & Pragmatic intent mostly preserved; minor loss of nuance
    that does not affect overall meaning. \\
3 & Core pragmatic intent present but cultural or tonal nuance
    is lost. \\
2 & Pragmatic intent partially lost; translation feels overly
    literal. \\
1 & Pragmatic intent completely lost; translation is misleading
    or pragmatically vacuous. \\
\bottomrule
\end{tabular}
\caption{Rubric used for human evaluation of translation outputs.
Annotators assessed pragmatic fidelity rather than surface fluency.}
\label{tab:rubric}
\end{table}

Annotators were explicitly instructed not to majorly penalise grammatical
imperfections if the pragmatic meaning was preserved, and not to reward
fluent outputs that missed the implied meaning. Each translation
rated independently without access to scores from other models or
annotators.
\subsection{COMET--Human Score Divergence Examples}
\label{subsec:divergence}

We present four translation outputs where COMET assigns a high score
but human raters assign 1 or 2 out of 5. These cases illustrate why
automatic metrics are sometimes insufficient for evaluating pragmatic translation
quality. COMET rewards lexical and structural similarity to the
reference but cannot detect pragmatically inverted or temporally
misanchored outputs.

\paragraph{Example 1: Deictic inversion (Malayalam, Deixis)}

\begin{mdframed}[
  topline=true, bottomline=true, rightline=true, leftline=true,
  linewidth=0.4pt, linecolor=black, backgroundcolor=gray!4,
  innerleftmargin=8pt, innerrightmargin=8pt,
  innertopmargin=5pt, innerbottommargin=5pt, roundcorner=2pt
]
\small
\textbf{Source:} \textit{``Njan ippozhum stationinte pinnile gate-il
aanu. \textbf{Ivide varenda}; nee avide thanne nilkku, njan athu
vannu cherkam.''}\\[3pt]
\textbf{Gloss:} ``I am still at the back gate of the station.
\textbf{Don't come here}; stay where you are, I'll come there.''\\[3pt]
\textbf{Model output (Qwen3-32B):}
``I am still at the station. \textbf{Come here}; you stand there,
I will come there.'' \\ [3pt]
\textbf{COMET:} 0.78 \quad \textbf{Human:} 1/5\\ [3pt]
\textit{The model inverts \textbf{ivide varenda} (don't come here)
to ``Come here'', producing a navigation instruction that is the
exact opposite of the source. COMET scores the output highly because
most content words are present; a human immediately recognises the
instruction is backwards.}
\end{mdframed}

\paragraph{Example 2: Subject-object inversion in implicature
(Punjabi, Implicature)}

\begin{mdframed}[
  topline=true, bottomline=true, rightline=true, leftline=true,
  linewidth=0.4pt, linecolor=black, backgroundcolor=gray!4,
  innerleftmargin=8pt, innerrightmargin=8pt,
  innertopmargin=5pt, innerbottommargin=5pt, roundcorner=2pt
]
\small
\textbf{Source:} \textit{``Surinder di salah leh ke na baithna,
uh tah \textbf{kath da ullu} hai.''}
\textbf{Gloss:} ``Don't take Surinder's advice;
\textbf{he is a complete fool}.''\\[3pt]
\textbf{Model output (Sarvam-105B):}
``\textbf{He made a fool of himself} by not listening to Surinder's
advice.'' \\ [3pt]
\textbf{COMET:} 0.76 \quad \textbf{Human:} 1/5\\ [3pt]
\textit{\textbf{Kath da ullu} is a common Punjabi meaning
``a complete fool.'' Sarvam mistranslates the subject of the insult
--- the output implies the speaker is foolish for ignoring Surinder,
whereas the source calls Surinder the fool. The pragmatic meaning is
exactly inverted.}
\end{mdframed}
\newpage
\paragraph{Example 3: Temporal deixis chain failure
(Punjabi, Deixis)}

\begin{mdframed}[
  topline=true, bottomline=true, rightline=true, leftline=true,
  linewidth=0.4pt, linecolor=black, backgroundcolor=gray!4,
  innerleftmargin=8pt, innerrightmargin=8pt,
  innertopmargin=5pt, innerbottommargin=5pt, roundcorner=2pt
]
\small
\textbf{Source:} \textit{``Seema ne Mangalvaar nu mainu dassya ki
Priya ne pichhle Somvaar nu unhu keha si, `pehla mera interview
\textbf{parso} si...'\,''}\\ [3pt]
\textbf{Gloss:} ``On Tuesday, Seema told me that Priya had told her
last Monday, `My interview was originally on \textbf{Wednesday}
[the day after tomorrow relative to that Monday]...'\,''\\ [3pt]
\textbf{Model output (Sarvam-105B):}
``My interview was the \textbf{day after tomorrow}...'' \\[3pt]
\textbf{COMET:} 0.86 \quad \textbf{Human:} 2/5 \\ [3pt]
\textit{The word \textbf{parso} means the day two steps from today--- past or future depending on context. The model fails to anchor it to
the correct speech time (last Monday), producing a future reference
where a past reference is required. COMET scores the output 0.86
because the words are nearly identical to the reference.}
\end{mdframed}

\paragraph{Example 4: Completed vs ongoing action deixis
(Tamil, Deixis)}

\begin{mdframed}[
  topline=true, bottomline=true, rightline=true, leftline=true,
  linewidth=0.4pt, linecolor=black, backgroundcolor=gray!4,
  innerleftmargin=8pt, innerrightmargin=8pt,
  innertopmargin=5pt, innerbottommargin=5pt, roundcorner=2pt
]
\small
\textbf{Source:} \textit{``Shanthi, unnoda friends ellaarum
\textbf{vanthutanga}, veliya saaptutu irukkanga, neeyum poi senthum
saappudu.''}\\[3pt]
\textbf{Gloss:} ``Shanthi, all your friends \textbf{have come} [already],
they are eating outside, you also go and join them.''\\ [3pt]
\textbf{Model output (Qwen3-32B):}
``Shanthi, all your friends \textbf{are coming}, you are also going
to join them.'' \\ [3pt]
\textbf{COMET:} 0.65 \quad \textbf{Human:} 1/5 \\ [3pt]
\textit{\textbf{Vanthutanga} is a perfective form indicating completed
action --- the friends have already arrived. The model translates it
as an ongoing future event, losing the urgency that the friends are
already there waiting. The deictic anchor shifts from present to future.}
\end{mdframed}


\subsection{Error Analysis}
\label{subsec:error_analysis}

We present five representative failure cases, one per phenomenon. Each case illustrates a distinct failure
mode that recurs across the benchmark.

\paragraph{Example 1: Face-saving invitation read literally
(Tamil, Implicature, MCQ)}

\begin{mdframed}[topline=true,bottomline=true,rightline=true,leftline=true,
  linewidth=0.4pt,linecolor=black,backgroundcolor=gray!4,
  innerleftmargin=8pt,innerrightmargin=8pt,
  innertopmargin=5pt,innerbottommargin=5pt,roundcorner=2pt]
\small
\textbf{Context:} \textit{``Veetukku vantha virunthaaligal purapidrapo
\textbf{kai nenachittu ponganu} amma sonnanga.''}\\
\textbf{Gloss:} ``As the guests who had come home were leaving, mother
told them: \textbf{wet your hands and go}.''\\[4pt]
\textbf{Question:} What did mother tell the guests?\\[3pt]
\begin{tabular}{ll}
A. Kai kaluva sonnanga (Told them to wash their hands) \\
\textbf{B. Inga saapida sonnanga}  \textbf{(Invited them to eat here)} \\
C. Veetukku poi saapida sonnanga  (Told them to go \\ home and eat) \\
D. Kulichutu poga sonnanga  (Told them to bathe \\ before leaving) \\
\end{tabular}\\[4pt]
\begin{tabular}{lll}
\textbf{Model} & \textbf{Prediction} & \textbf{Gold} \\
\hline
Llama-3.1-8B  & C \ding{55} & \textbf{B} \\
Gemma-4-31B   & C \ding{55} & \textbf{B} \\
Qwen3-32B     & C \ding{55} & \textbf{B} \\
Sarvam-105B   & A \ding{55} & \textbf{B} \\
Command-A & A \ding{55} & \textbf{B} \\
\end{tabular}\\[4pt]
\textit{\textbf{Kai nenachittu po} (``wet your hands and go'') is a
traditional Tamil farewell invitation to stay and eat --- a face-saving
formula that allows guests to decline gracefully. Llama, Gemma, and Qwen
interpret it as telling guests to go home and eat (C); Sarvam reads it
literally as a hygiene instruction (A). No model recovers the implied
invitation to stay (B).}
\end{mdframed}

\paragraph{Example 2: Referent confusion in family dialogue
(Tamil, Deixis, MCQ)}

\begin{mdframed}[topline=true,bottomline=true,rightline=true,leftline=true,
  linewidth=0.4pt,linecolor=black,backgroundcolor=gray!4,
  innerleftmargin=8pt,innerrightmargin=8pt,
  innertopmargin=5pt,innerbottommargin=5pt,roundcorner=2pt]
\small
\textbf{Context:} \textit{``Unakku thevayaana thuni mani ellam eduthu
vechachanu amma \textbf{unkitta} kekka sonnanga. Ne eduthuvechuttiyanu
en \textbf{thangachi kitta} na ketten.''}\\
\textbf{Gloss:} ``Amma asked me to check with you [unkitta] whether
you had taken all the cloth pieces you needed. I asked my younger
sister [thangachi] whether she had taken them.''\\[4pt]
\textbf{Question:} Who does the speaker refer to with `\textbf{unakku}'?\\[3pt]
\begin{tabular}{ll}
A. Appa & B. Amma \\
\textbf{C. Thangachi} & D. Akka \\
\end{tabular}\\[4pt]
\begin{tabular}{lll}
\textbf{Model} & \textbf{Prediction} & \textbf{Gold} \\
\hline
Llama-3.1-8B  & B (Amma) \ding{55}  & \textbf{C (Thangachi)} \\
Gemma-4-31B   & C \ding{52}        & \textbf{C (Thangachi)} \\
Qwen3-32B     & B (Amma) \ding{55}  & \textbf{C (Thangachi)} \\
Sarvam-105B   & C \ding{52}        & \textbf{C (Thangachi)} \\
Command-A & B (Amma) \ding{55} & \textbf{C (Thangachi)} \\
\end{tabular}\\[4pt]
\textit{The utterance involves three participants (amma, speaker, thangachi)
across two turns. Llama and Qwen conflate amma and thangachi, failing to
track the referential chain across the two embedded clauses.}
\end{mdframed}

\paragraph{Example 3: Hyperbolic repetition read literally
(Punjabi, Speech Acts, NLI)}

\begin{mdframed}[topline=true,bottomline=true,rightline=true,leftline=true,
  linewidth=0.4pt,linecolor=black,backgroundcolor=gray!4,
  innerleftmargin=8pt,innerrightmargin=8pt,
  innertopmargin=5pt,innerbottommargin=5pt,roundcorner=2pt]
\small
\textbf{Premise:} \textit{``Papa ne naraaz hoke Raj nu keha, `Main tenu
\textbf{100 vari} keha, ghar vich ball na kheliya kar.'\,''}\\
\textbf{Gloss:} ``Papa angrily told Raj, `I have told you
\textbf{100 times}, don't play ball inside the house.'\,''\\[4pt]
\textbf{Hypothesis:} Papa has already told Raj this same thing
99 times before.\\[4pt]
\begin{tabular}{lll}
\textbf{Model} & \textbf{Prediction} & \textbf{Gold} \\
\hline
Llama-3.1-8B  & Entailment \ding{55}    & \textbf{Contradiction} \\
Gemma-4-31B   & Contradiction \ding{52} & \textbf{Contradiction} \\
Qwen3-32B     & Entailment \ding{55}    & \textbf{Contradiction} \\
Sarvam-105B   & Neutral \ding{55}       & \textbf{Contradiction} \\
Command-A & Entailment \ding{55} & \textbf{Contradiction} \\
\end{tabular}\\[4pt]
\textit{``100 vari'' is a hyperbolic speech act expressing frustration,
not a literal count. The hypothesis that papa said it exactly 99 times
before contradicts the pragmatic meaning. Llama and Qwen treat it as a
factual claim and predict Entailment.}
\end{mdframed}
\newpage
\paragraph{Example 4: Veiled criticism read as neutral observation
(Punjabi, Social Pragmatics, NLI)}

\begin{mdframed}[topline=true,bottomline=true,rightline=true,leftline=true,
  linewidth=0.4pt,linecolor=black,backgroundcolor=gray!4,
  innerleftmargin=8pt,innerrightmargin=8pt,
  innertopmargin=5pt,innerbottommargin=5pt,roundcorner=2pt]
\small
\textbf{Premise:} \textit{``Kal jado mainu meri maasi mili, unha ne
keha, `Tu taan bada \textbf{kamzor lag reha hai.}'\,''}\\
\textbf{Gloss:} ``Yesterday when I met my aunt, she said,
`You look quite thin/weak.'\,''\\[4pt]
\textbf{Hypothesis:} The aunt was mocking the speaker's health.\\[4pt]
\begin{tabular}{lll}
\textbf{Model} & \textbf{Prediction} & \textbf{Gold} \\
\hline
Llama-3.1-8B  & Entailment \ding{55} & \textbf{Contradiction} \\
Gemma-4-31B   & Neutral \ding{55}    & \textbf{Contradiction} \\
Qwen3-32B     & Neutral \ding{55}    & \textbf{Contradiction} \\
Sarvam-105B   & Neutral \ding{55}    & \textbf{Contradiction} \\
Command-A & Neutral \ding{55} & \textbf{Contradiction} \\
\end{tabular}\\[4pt]
\textit{In Punjabi culture, an elder commenting on a younger person's
appearance (``you look thin'') is a conventional expression of concern and
care, not mockery. The correct label is Contradiction --- the aunt is
being caring, not sarcastic. All four models miss this cultural norm.}
\end{mdframed}

\paragraph{Example 5: Topically related but discourse-incoherent
(Tamil, Coherence, MCQ)}

\begin{mdframed}[topline=true,bottomline=true,rightline=true,leftline=true,
  linewidth=0.4pt,linecolor=black,backgroundcolor=gray!4,
  innerleftmargin=8pt,innerrightmargin=8pt,
  innertopmargin=5pt,innerbottommargin=5pt,roundcorner=2pt]
\small
\textbf{Context:} \textit{``Tamilnadu, India la irukurathulaaye south la
irukkura state. Idhu areavula 10-avathu periya state, janathila 6-avathu
periyadhu. Tamilnadula la iruka makkal tamil pesuvaanga...''}\\
\textbf{Gloss:} ``Tamil Nadu is a southern state of India. It is the
10th largest by area and 6th most populous. People in Tamil Nadu speak
Tamil...''\\[4pt]
\textbf{Question:} What is the most appropriate next sentence?\\[3pt]
\begin{tabular}{lp{0.7\columnwidth}}
A. & Tamilnadukku naraya sutrula payanigal varuvaanga (Many tourists visit) \\
\textbf{B.} & \textbf{Chennai dhan thalainagaram (Chennai is the capital)} \\
C. & Tamil classica solappadum Thirukkural... (literary reference) \\
D. & Tamil cinema industryah kollywoodnu solluvaanga (Kollywood) \\
\end{tabular}\\[4pt]
\begin{tabular}{lll}
\textbf{Model} & \textbf{Prediction} & \textbf{Gold} \\
\hline
Llama-3.1-8B  & A \ding{55} & \textbf{B} \\
Gemma-4-31B   & B \ding{52} & \textbf{B} \\
Qwen3-32B     & C \ding{55} & \textbf{B} \\
Sarvam-105B   & D \ding{55} & \textbf{B} \\
Command-A & B \ding{52} & \textbf{B} \\
\end{tabular}\\[4pt]
\textit{The paragraph introduces factual information about Tamil Nadu
systematically (area, population, language). The next logical fact is the
capital city. Options A, C, D are all topically related to Tamil Nadu but
break the structured informational progression. Three models pick
topically plausible but discourse-incoherent continuations.}
\end{mdframed}